\documentclass[letterpaper, 10 pt, conference]{ieeeconf}  

\IEEEoverridecommandlockouts                              

\usepackage{times} 
\usepackage{amsmath} 
\usepackage{amssymb}  
\usepackage{url}
\usepackage{latexsym}
\usepackage{graphicx}	
\usepackage{comment}
\usepackage{threeparttable} 
\usepackage{scalefnt}
\usepackage{multicol} 
\usepackage{wrapfig}  
\usepackage{algorithm}  
\usepackage{algpseudocode}
\usepackage{here}  
\usepackage{autobreak}  
\usepackage{breqn}
\usepackage{bm}

\makeatletter  
\newcommand\fs@spaceruled{\def\@fs@cfont{\bfseries}\let\@fs@capt\floatc@ruled
  \def\@fs@pre{\vspace{0.5\baselineskip}\hrule height.2pt depth0pt \kern2pt}%
  \def\@fs@post{\kern2pt\hrule\relax}%
  \def\@fs@mid{\kern2pt\hrule\kern2pt}%
  \let\@fs@iftopcapt\iftrue}
\makeatother

\title{\LARGE \bf
Learning Bidirectional Translation between Descriptions and Actions with Small Paired Data
}

\author{Minori Toyoda*$^{1}$, Kanata Suzuki*$^{1,2}$,Yoshihiko Hayashi$^{1}$, and Tetsuya Ogata$^{1,3,4}$
\thanks{
* These authors have contributed equally to this work.
$^{1}$Minori Toyoda, $^{1,2}$Kanata Suzuki, $^{1}$Yoshihiko Hayashi and $^{1,3,4}$Tetsuya Ogata are affiliated with Faculty of Science and Engineering, Waseda University, Tokyo 169-8050, Japan. 
$^{1,2}$Kanata Suzuki is also at Artificial Intelligence Laboratories, Fujitsu Limited., Kanagawa 211-8588, Japan. 
$^{1,3,4}$Tetsuya Ogata is also with the Waseda Research Institute for Science and
Engineering (WISE) at Waseda University, Tokyo 169-8555, Japan, and the National Institute of Advanced Industrial Science and Technology, Tokyo 100-8921, Japan.
E-mail: {\tt\small minori-toyoda@fuji.waseda.jp}
.}
}

\begin{document}

\maketitle
\thispagestyle{empty}
\pagestyle{empty}


\begin{abstract}
This study achieved bidirectional translation between descriptions and actions using small paired data from different modalities.
The ability to mutually generate descriptions and actions is essential for robots to collaborate with humans in their daily lives, which generally requires a large dataset that maintains comprehensive pairs of both modality data.
However, a paired dataset is expensive to construct and difficult to collect. 
To address this issue, this study proposes a two-stage training method for bidirectional translation. 
In the proposed method, we train recurrent autoencoders (RAEs) for descriptions and actions with a large amount of non-paired data. 
Then, we fine-tune the entire model to bind their intermediate representations using small paired data. 
Because the data used for pre-training do not require pairing, behavior-only data or a large language corpus can be used. We experimentally evaluated our method using a paired dataset consisting of motion-captured actions and descriptions.
The results showed that our method performed well, even when the amount of paired data to train was small. The visualization of the intermediate representations of each RAE showed that similar actions were encoded in a clustered position
and the corresponding feature vectors were well aligned.

\end{abstract}


\section{Introduction}
Robots
that collaborate with humans are required to have the ability to generate actions from language instructions and explain their actions.
If this is the case, we would not need to devise a complicated system to command a robot.
Besides, a human user could easily understand the behavior of the robot.
To achieve this ability, the robot needs to associate real-world objects with linguistic expressions, and a comprehensive paired dataset is generally required to acquire that relationship.
In particular, the machine learning-based approaches~\cite{Yamada2018_PRAE}\cite{Toyoda2021_rPRAE}
require the preparation of a large paired dataset of descriptions and actions.
However, collecting and captioning behavioral data is costly, making it difficult to collect large amounts of paired data.

\begin{figure}[t]
\centering
\includegraphics[width=\columnwidth]{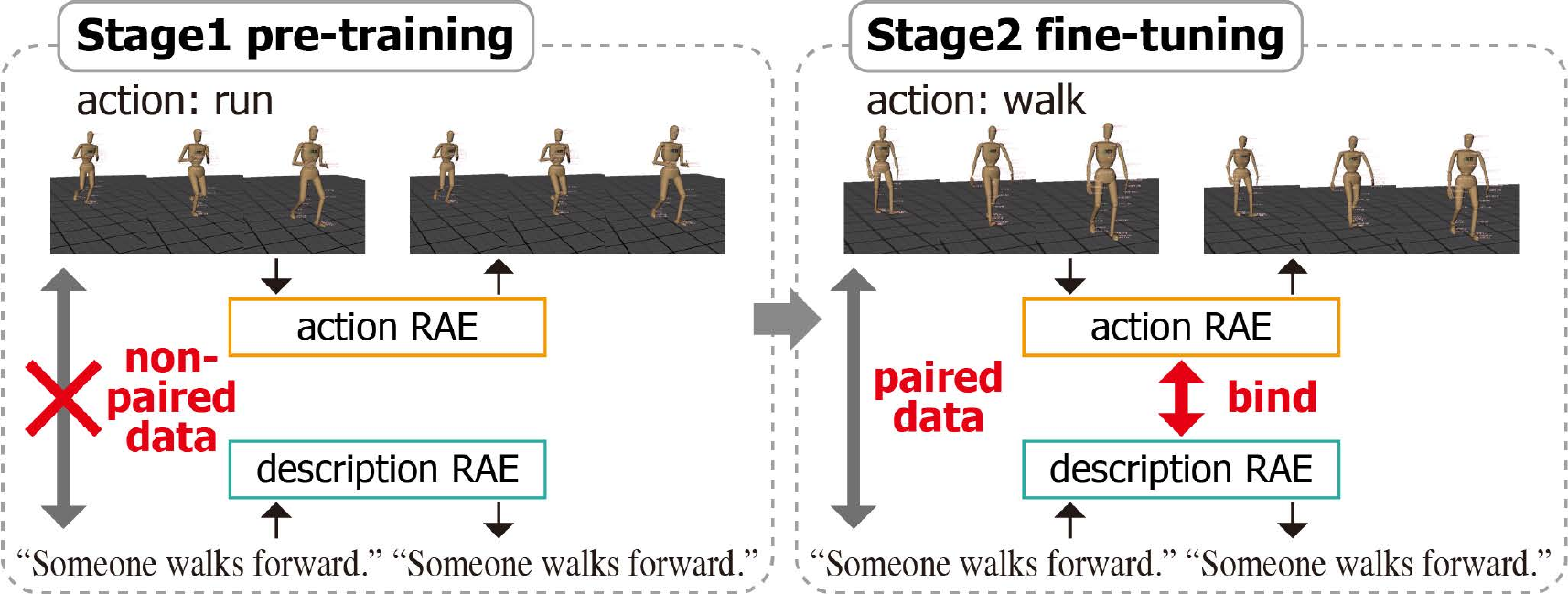}
\caption{
Overview of this study. [Stage1] First, we pre-train two recurrent autoencoders (RAEs) with large-scaled 
non-paired
data for linguistic descriptions and actions.
[Stage2] Subsequently, we fine-tune the entire model by making closer the intermediate representations of both RAEs with small-scale paired data. A binding loss is employed in training.
}
\label{fig:1}
\end{figure}

Several joint learning methods for language and motion have been proposed to address these issues. 
Using context-independent word embeddings derived from a large corpus,
\cite{Toyoda2021_rPRAE} and \cite{Matthews2019} dealt with unseen words that were not included in the training data.
In other studies, stepwise learning methods were proposed to integrate multiple models from different modalities~\cite{Qiu2020pre-train}\cite{zellers2021piglet}.
These previous studies demonstrated that pre-trained models trained using a large dataset can be effectively exploited by being fine-tuned using task-oriented data.
Motivated by these works, we pre-train a model in each modality (language and motion) and fine-tune the entire bidirectional translation model using small paired data.

The current study proposes a two-stage training method that does not require a comprehensive paired dataset for bidirectional translation between descriptions and actions. An overview of this study is presented in Fig.~\ref{fig:1}. First, we pre-train two recurrent autoencoders (RAEs) for descriptions and actions separately with 
non-paired
data, and then fine-tune the entire model by binding their intermediate representations using small paired data. Because the data used for pre-training do not need to be paired, 
behavior data and a large language corpus can be independently used for training those RAEs.
Our experiments used a paired dataset of motion capture and reference caption to perform mutual generation of descriptions and actions.
We performed quantitative and qualitative evaluations using large paired datasets, and an analysis of model behavior.

\begin{figure*}[tb]
\centering
\includegraphics[width=17cm]{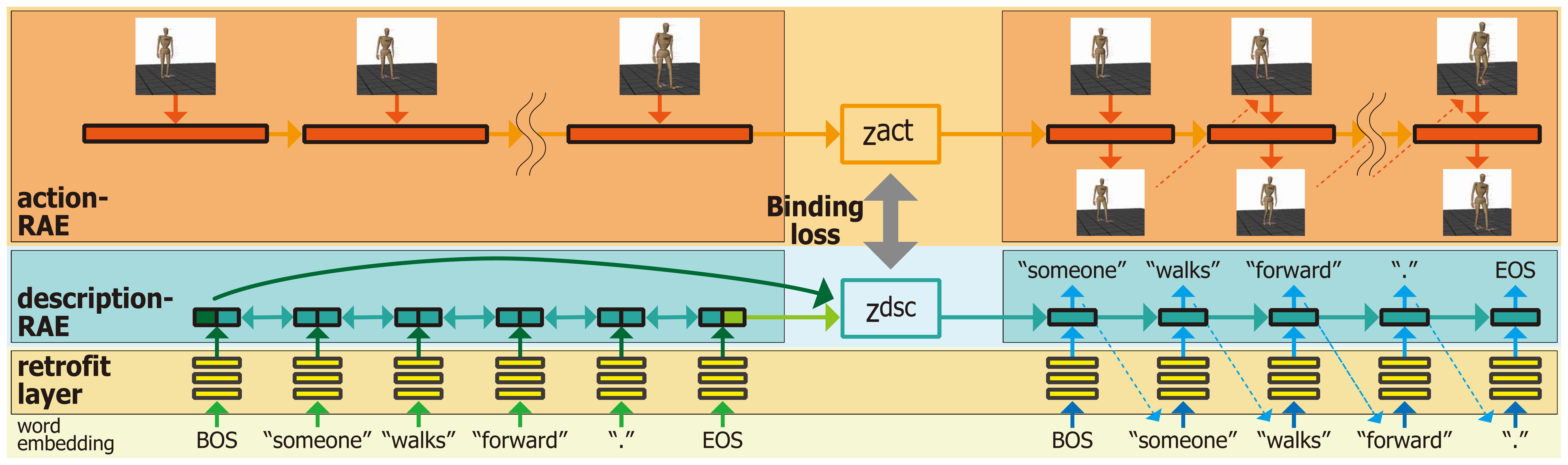}
\caption{retrofitted Paired Recurrent Autoencoder (rPRAE).}
\label{fig:2}
\end{figure*}

\section{Related Work}
The present work aims at the bidirectional translation of descriptions and actions without the need for comprehensive paired data.
The efficient use of non-paired and/or small data would be vital in this regard.


\subsection{Utilization of Pre-trained Model in Translation Tasks}
Many studies have proposed methods for training translation/transformation models with limited data resources.
One solution is to use a pre-trained model in the source/target modality.
In the area of natural language processing, several combinations of source/target modalities have been worked out by using pre-trained models~\cite{Qiu2020pre-train}\cite{ruan2022video-language}.
The quality and quantity of the cross-modal data have been central issues to tackle in these studies.
Using a pre-trained model in the source/target side could partly address the issue, particularly the quantity issue.
In fact, large pre-trained language models, such as BERT~\cite{bert}, have significantly contributed to the improvement in the cross-modal task performances.


Several studies have been conducted on bidirectional translation tasks.
With regard to the bidirectional text--to/from--image generation task, there are some studies that apply pre-trained models trained with large-scale data~\cite{huang2021unifying}. 
During the conversion between symbolic actions/states and texts in a 3D simulator environment~\cite{zellers2021piglet},
pre-training on large non-paired data has shown an improved performance in zero-shot settings.
For description--from/to--action translation, \cite{Toyoda2021_rPRAE} realized appropriate action generation that accepts unseen words not included in the dataset.
They successfully retrofitted the pre-trained word embeddings into multimodal representations incorporating the action modality.

Creating a pre-trained model with a single modality is relatively easy, particularly by applying self-supervised learning.
The effective use of a pre-trained unimodal model could address the issue of lacking paired data in translation/transformation, provided a proper integration of the representations in a particular modality is achieved.

\subsection{Integration of Multimodal Representations}
When data resources are limited, the method of integrating the two modalities, that is, accurate connection of the pre-trained models, is important.
Multimodal representations can be categorized into two types according to the way they are acquired:
joint and coordinated representations~\cite{baltruvsaitis2018multimodal}.

A joint representation is obtained by mapping all modalities into the same representation space~\cite{zellers2021piglet}\cite{NODA2014721}.
In object manipulation and bell-ringing tasks, \cite{NODA2014721} obtained multimodal representations by concatenating the extracted representations such as images and sound signals with joint angles and passing them through an autoencoder.
In this study, the feature extractor needs to be trained from scratch, which makes it difficult to train with a small amount of paired data.
This requirement of large paired data is a general issue of joint representation~\cite{baltruvsaitis2018multimodal}.

A coordinated representation is obtained by enforcing similarity constraints and is often used in translation tasks~\cite{baltruvsaitis2018multimodal}.
The present study also adopts this method, following other studies of bidirectional translation between descriptions and actions~\cite{Yamada2018_PRAE}\cite{Toyoda2021_rPRAE}.
A bi-modal autoencoder model has been proposed for the acquisition of multimodal representations of vision and language~\cite{hasegawa-etal-2017-incorporating}.
Appropriate multimodal word representations were obtained using pre-trained word embeddings and visual features, and learning to bring the autoencoder's intermediate representations of word representations and visual features closer together.
This constraint allowed us to train the entire model end-to-end at a lower computational cost than that of the joint representation.

To achieve efficient bidirectional translation between descriptions and actions with small amounts of data, we pre-trained the language and action models separately and then fine-tuned them by imposing a learning constraint that binds the intermediate representations of both modalities.

\section{Method}
\subsection{Model Overview}
In this study, we use a retrofitted paired recurrent autoencoder (rPRAE~\cite{Toyoda2021_rPRAE}) to perform bidirectional translation between descriptions and actions. 
An overview of the model is presented in Fig.~\ref{fig:2}.
rPRAE comprises two RAEs for action and description, and a retrofit layer, which is a nonlinear layer that transforms word embeddings. Each RAE has a sequence-to-sequence model structure comprising encoder and decoder parts, and is trained to output the same data as the input
time-series 
data. 
The encoder extracts low-dimensional feature vectors of the input data. In the rPRAE, the extracted feature vectors are input to the decoder of the other RAE to perform a bidirectional translation. In this paper, we refer to these feature vectors as intermediate representations.

In the action RAE, the joint angles and position/rotation of the body are used as the input and output of the model\footnote{Although the dataset used in this work excludes environmental information such as vision, representations that reflect the information necessary for dynamic description/action generation can be obtained by concatenating it with motion signals and inputting it to the action RAE as demonstrated in~\cite{Toyoda2021_rPRAE}, provided that the experimental setting is simple.}.
The loss function $L_{\rm act}$ of the action RAE is defined based on the squared error, as follows:
\begin{align}
    L_{\rm act} = \sum ^{T_a-1}_{t=1} \left\|\bm{j}_{t+1} - \bm{\widehat{j}}_{t+1}\right\| ^2_2,
\end{align}
where $T_a$ denotes the length of the input motion sequence, $\bm{j}$ denotes the predicted motion signal, and $\bm{\hat{j}}$ denotes the teaching signal.

In contrast, in the description RAE, the pre-trained word embedding is transformed by the nonlinear layer and input to the encoder. The decoder outputs the probability of occurrence of words in the vocabulary. The loss function of the description RAE, $L_{\rm dsc}$, uses cross entropy and is defined as follows:
\begin{align}
    L_{\rm dsc} = - \sum^{T_d-1} _{t=1} \sum^W_w \bm{x}_{t+1} \left(w\right) {\rm log} \bm{y}_{t} \left(w\right),
\end{align}
where $T_d$ is the length of the input description sequence, $W$ is the vocabulary size, $\bm{x}$ is the word embedding input to the model, and $\bm{y}$ is the output of the model.

In addition to the reconstruction errors of RAEs, the rPRAE learns to approximate the intermediate representations output by the encoders. By binding the intermediate representations of both RAEs, bidirectional translation can be performed by inputting the feature vectors extracted by one RAE encoder into the decoder of the other RAE. The loss function $L_{\rm bnd}$ for the intermediate representation is defined as follows:
\begin{eqnarray}
    L_{\rm bnd} = \sum _i^B \psi \left(\bm{z}_{i}^{\rm act}, \bm{z}_{i}^{\rm dsc} \right) \hspace{140pt} \nonumber \\
    +\sum _i^B \sum _{j\ne i} {\rm max} \left\lbrace 0, \Delta + \psi \left(\bm{z}_{i}^{\rm act}, \bm{z}_{i}^{\rm dsc} \right) - \psi \left(\bm{z}_{i}^{\rm act}, \bm{z}_{j}^{\rm dsc} \right)\right\rbrace, \nonumber
\end{eqnarray}
\vspace{-17pt}
\begin{eqnarray}
\end{eqnarray}
where $B$ is the batch size, $\left\{\bm{z}_i^{dsc} | 1\leq i\leq B\right\}$ represents the intermediate representation of the description RAE, and $\left\{\bm{z}_i^{act} | 1\leq i\leq B\right\}$ is the intermediate representation of the action RAE. In this study,
$\psi$ is the Euclidean distance between the feature vectors, but other losses can be used as well. 
$\Delta$ denotes the margin. The first term brings the corresponding intermediate representations closer, and the second term moves the non-corresponding ones further. This enables integrated learning to map intermediate representations of descriptions and actions in the feature space.

For the description RAE to extract intermediate representations similar to that of the action RAE, it is desirable that the input of the description RAE encoder also contains motion information. We also train the retrofit layer to output word vectors that include information on the action. Because of the alternating parameter updates of the RAE and retrofit layer described in the following subsection, the output of the retrofit layer has similar vector representations of words for each action meaning.

\subsection{Two-Stage Training with Small Paired Data}
This study proposes a two-stage training method for rPRAE that does not require a large amount of paired data. As shown in Algorithm 1, the proposed method comprises two training steps.

\textbf{Training stage 1:}
First, we pre-train the RAEs using only the reconstruction errors shown in equations (1) and (2). In this training stage, only non-paired data $X^{\rm dsc}_{\rm NP}, X^{\rm act}_{\rm NP}$ are required (line 3 in Algorithm 1). The loss function in training stage 1 $L_{\rm S1}$ is defined as follows:
\begin{align}
    \label{eq:loss_pre}
    L_{\rm S1} = L_{\rm dsc} + L_{\rm act},
\end{align}
Using independent loss functions in the RAEs, each model is trained to extract action- and description-specific representations. In addition, it is possible to use non-paired datasets for training, which allows us to leverage uncaptioned actions and a large language corpus.

\floatstyle{spaceruled}
\restylefloat{algorithm}
\begin{algorithm}[bt] 
    \caption{Training procedures for the rPRAE}
    \label{alg:training}
    \begin{algorithmic}[1]
        \Require $X^{\rm dsc}_{\rm NP}, X^{\rm act}_{\rm NP}:$ non-paired dataset
        \Require $N_1, N_2, n_{\rm ch}:$ hyperparameters
        \State randomly initialize learnable parameters: $\theta_{\rm rae}, \theta_{\rm ret}$
        \For{$i=1$ to $N_1$} \Comment{Training of stage 1}
            \State Sample non-paired data $x^{\rm dsc}_{\rm NP}, x^{\rm act}_{\rm NP}$ from $X^{\rm dsc}_{\rm NP}, X^{\rm act}_{\rm NP}$
            \State Calculate $L_{\rm dsc}, L_{\rm act}$ by forward-path
            \State Compute total loss: $L_{\rm S1} \Leftarrow L_{\rm dsc} + L_{\rm act}$
            \If{$int(i/n_{ch})\%2=1$}
                \State Update $\theta_{\rm rae}$ by gradient method w.r.t. $L_{\rm S1}$
            \Else
                \State Update $\theta_{\rm ret}$ by gradient method w.r.t. $L_{\rm S1}$
            \EndIf
        \EndFor
        \State make small paired dataset $X^{\rm dsc}_{\rm P}, X^{\rm act}_{\rm P}$ from $X^{\rm dsc}_{\rm NP}, X^{\rm act}_{\rm NP}$
        \For{$i=1$ to $N_2$}
            \Comment{Training of stage 2}
            \State Sample paired batch $\left\{ x^{\rm dsc}_{\rm P}, x^{\rm act}_{\rm P}\right\}$ from $X^{\rm dsc}_{\rm P}, X^{\rm act}_{\rm P}$
            \State Calculate $L_{\rm dsc}, L_{\rm act}, L_{\rm bnd}$ by forward-path
            \State Compute total loss: $L_{\rm S2} \Leftarrow L_{\rm dsc} + L_{\rm act} +L_{\rm bnd}$
            \If{$int((i+N_1)/n_{ch})\%2=1$}
                \State Update $\theta_{\rm rae}$ by gradient method w.r.t. $L_{\rm S2}$
            \Else
                \State Update $\theta_{\rm ret}$ by gradient method w.r.t. $L_{\rm S2}$
            \EndIf
        \EndFor
    \end{algorithmic}
\end{algorithm}

\textbf{Training stage 2:}
After pre-training, we fine-tune the entire model using a small amount of paired data. Here, small paired data ${X^{\rm dsc}_{\rm P}, X^{\rm act}_{\rm P}}$ are obtained by captioning a portion of the non-paired data $X^{\rm dsc}_{\rm NP}, X^{\rm act}_{\rm NP}$ (line 12 in Algorithm 1). In addition to the reconstruction errors, the learning constraints in Equation (3) are applied to bring the intermediate representations of the RAEs closer together. The loss function in training stage 2 $L_{\rm S2}$ is defined as follows:
\begin{align}
    \label{eq:loss}
    L_{\rm S2} = L_{\rm dsc} + L_{\rm act} + L_{\rm bnd},
\end{align}
This allows the RAEs to retain their pre-trained representation while mapping each representation of the modality.

In both training stages 1 and 2, we update the parameters of RAEs $\theta_{\rm rae}$ and retrofit layer $\theta_{\rm ret}$ alternately for every certain number of $n_{\rm ch}$ (lines 6--10 and 17--21 in Algorithm 1). This method refers to ``chain-thaw~\cite{felbo2017using},'' which fine-tunes the parameters of the model one layer at a time to prevent overtraining of the retrofit layer due to an increase in training parameters. By training the layers separately, we obtain the word representations without overfitting to the small paired data while moderately limiting the expressive ability of each model.
Additionally, the retrofit layer can map the word embedding to a similar feature space for each action~\cite{Toyoda2021_rPRAE}.

Using the aforementioned two-stage training process, we obtain word embeddings that match the actions and update the parameters so that the feature space of each pre-trained modality is equalized with the small paired data.
The proposed method has a simple structure; however, it can be applied to a variety of data types.
The small paired data should be obtained in a similar setting to the non-paired data used in training stage 1 because training stage 2 requires binding the independent intermediate representations.
This issue should be investigated further in future studies.

\begin{figure}[tb]
\centering
\includegraphics[width=\columnwidth]{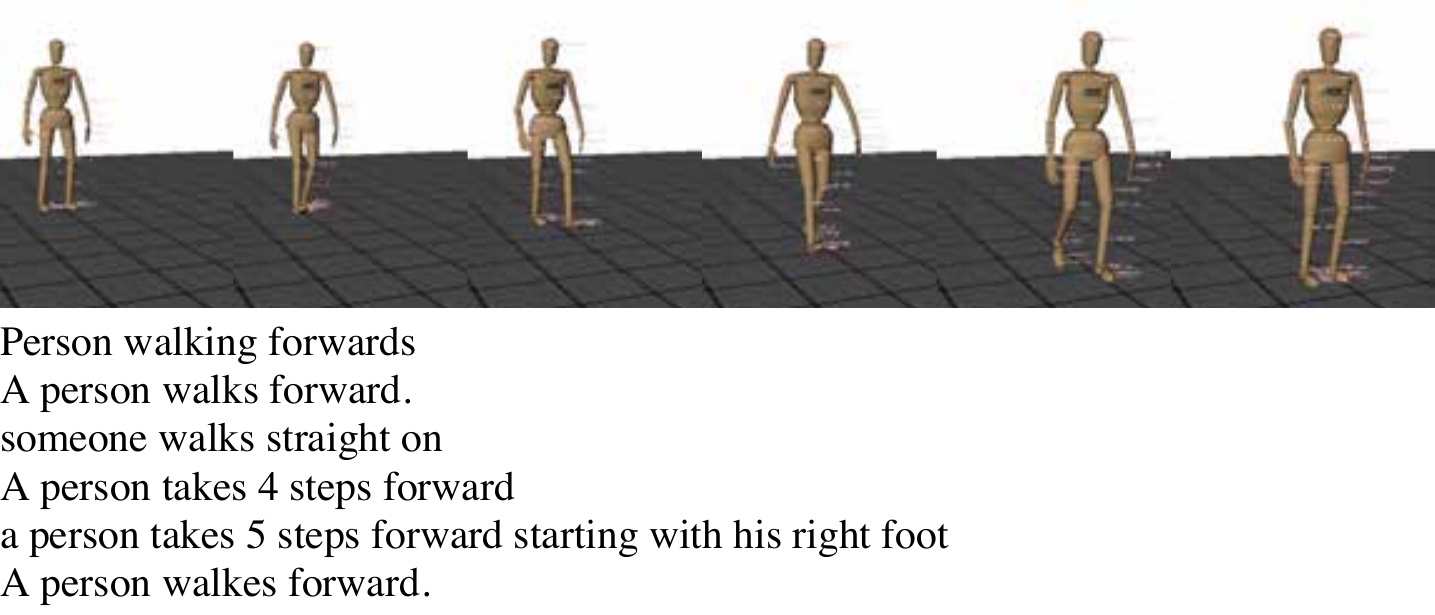}
\caption{Example of an action and descriptions in KIT dataset.}
\label{fig:3}
\end{figure}

\begin{figure}[tb]
\centering
\includegraphics[width=\columnwidth]{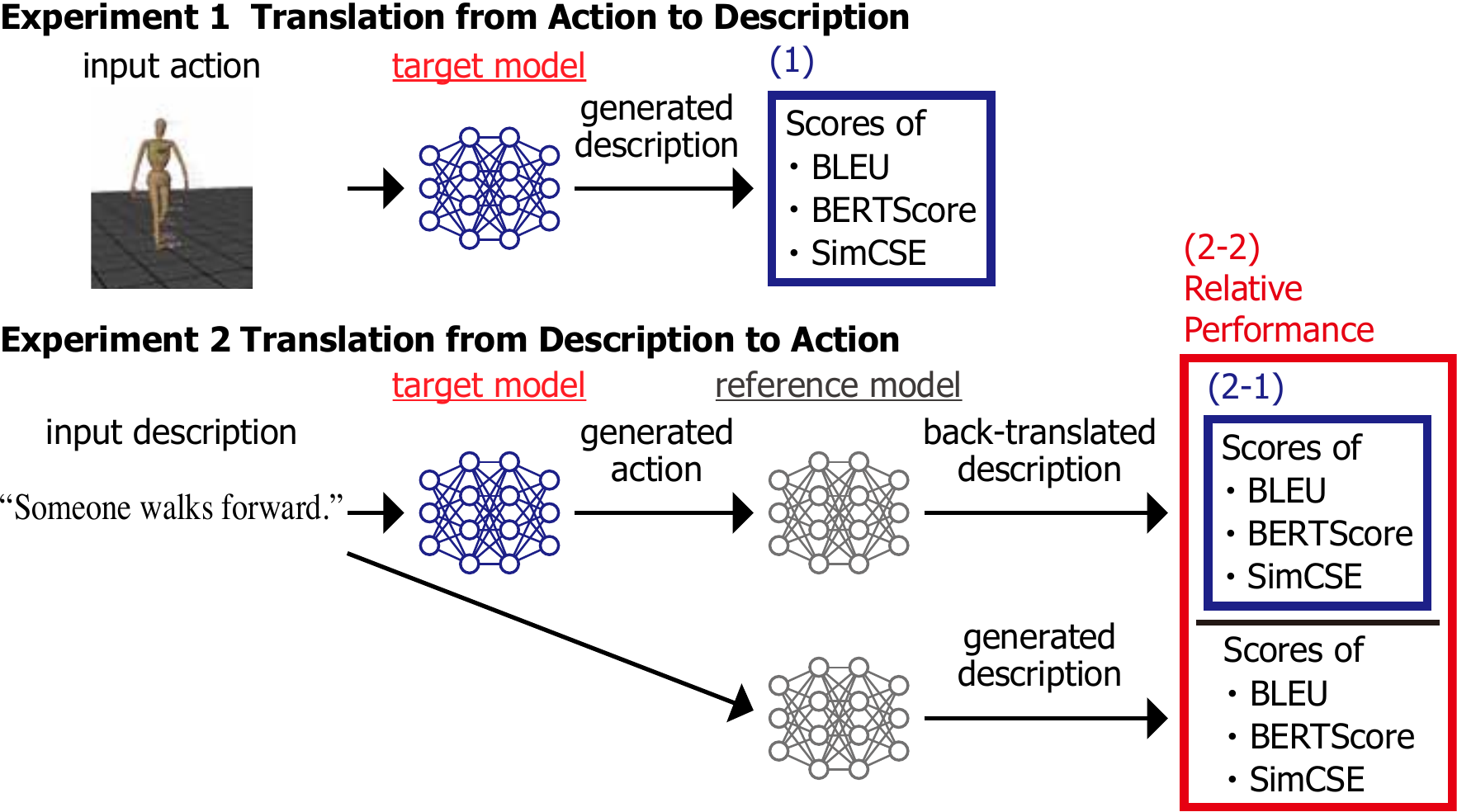}
\caption{Evaluation metrics in experiments 1 and 2. (1) and (2-1) indicate the scores of the generated descriptions in experiments 1 and 2, respectively. (2-2) shows the relative performance to the reference model.}
\label{fig:4}
\end{figure}

\section{Experiments}
\subsection{Dataset}
\label{sec:dataset}
In this study, we performed experiments on the mutual generation of descriptions and actions with a large paired dataset: the KIT Motion-Language Dataset~\cite{Plappert2016_KIT}, which comprises motion-captured actions and reference captions.
Examples of an action and descriptions are presented in Fig.~\ref{fig:3}. 
The KIT dataset contains 3911 motion sequences, each represented by an ordered sequence of 50-dimensional vectors: 44 joint angles of the whole body, three values of the body's root position, and three values of its rotation.
In the experiments, following the previous studies~\cite{Yamada2018_PRAE}\cite{Plappert2018}, each motion of 100--3000 steps was downsampled to $1/10^{th}$ of the sampling frequency\footnote{
We used only the paired data whose captions were formed with words equipped with the pre-trained embeddings. This selection of data resulted in a subset of 2721 actions and 5538 caption sentences. The vocabulary size is $W=1331$; the max sentence length in the number of words is 38.}. 
All the motion data values were normalized to the empirically determined range [-0.9, 0.9].
The KIT dataset also contains 6353 caption sentences with a vocabulary size of $W=1646$. Each sentence consisted of 5--44 words, including the BOS and EOS, at the beginning and end.
For each word, we extracted a 300-dimensional vector representation using a pre-trained Word2Vec~\cite{Mikolov2013_w2v}.
Further, 80\% of the available action sequences were used for training, 10\% for validation, and the remaining 10\% for evaluation (Table~\ref{tab:0}).
Because there were multiple captions for a single action, paired data were created for the number of captions.
Therefore, the total number of paired data for training was 4368.

\begin{table}[tb]
\caption{Data Division}
    \centering
    \begin{tabular}{c|cc}
    \hline
    & Number of Action & Number of Caption \\
    \hline
    \hline
    Train      & 2176 & 4368 \\
    Validation & 272  & 610 \\
    Test       & 273  & 560 \\
    \hline
    \end{tabular}
\label{tab:0}
\end{table}

\begin{table}[t]
    \centering
    \caption[Model Structures]
    {Structure of the models}
    \begin{tabular}{c|c}
        \hline
        Model         & Layer and Dimension \\
        \hline \hline
        action-RAE*  & Input@50 - LSTM@500 - $z^{act}$@500 - \\ 
                   & LSTM@(500-500-500) - Output@50 \\ 
        \hline
        description-RAE*  & Input@300 - BiLSTM@500 - $z^{dsc}$@500 - \\ 
                   & LSTM@500 - Output@1331 \\
        \hline 
        retrofit layer* & Input@300 - FC@(400-400-400) - Output@300 \\ 
        \hline 
    \end{tabular}
    \begin{flushleft}
        *) All layers have tanh activation, except for the transformation part of the intermediate representation $z^{act}$ and $z^{dsc}$.
    \end{flushleft}
\label{tab:1}
\end{table}

\begin{figure*}[tb]
\centering
\includegraphics[width=17cm]{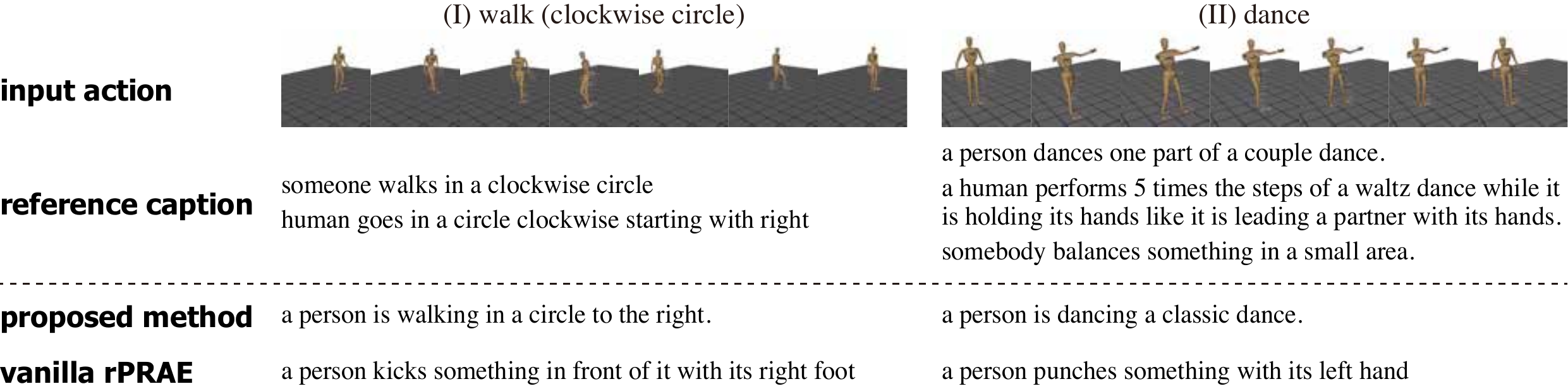}
\caption{Generated samples of descriptions. The pictures show the input actions, and each description below the broken line represents the generated description. The proposed model was pre-trained with 4368 non-paired data and fine-tuned with 500 paired data.
The vanilla rPRAE was trained with 500 paired data.}
\label{fig:5-1}
\end{figure*}

\subsection{Training Setup}
The configuration of each model is presented in Table~\ref{tab:1}. LSTMs are used for the encoder and decoder parts of each RAE, and only the encoder of the description RAE uses a bidirectional LSTM~\cite{bi-rnn}. Each LSTM consisted of one layer with 500 neurons, and only the decoder of the action RAE consisted of three LSTM layers. The retrofit layer consisted of three fully-connected (FC) layers with 400 neurons, and the input and output had 300 dimensions. The intermediate representation of each RAE has 500 dimensions. We used tanh as the activation function. The batch size $B$ was set to 80, and Adam~\cite{adam} was used as the optimizer. 
The number of training epochs was set to 200 for training stage 1 and 400 for training stage 2. 
We trained 4368 sequences (described in Section ~\ref{sec:dataset}); thus, 
when using the comprehensive training data in stage 1 and 500 paired items in stage 2, one epoch contains 55 and 7 iterations, respectively.
Therefore, the total number of iterations for stage 1 is $N_1=11000$ and that for stage 2 is  $N_2=2800$.
In both training stages 1 and 2, the parameters of the RAEs and retrofit layer were updated by alternating every $n_{\rm ch}=$100 updates. 
These parameters were determined after considering several training settings, and there was no particular sensitivity problem if the training epochs $N_1$ and $N_2$ were large enough.
We compared the performance of rPRAE without training stage 1 as the conventional method (vanilla rPRAE) with that of the proposed method, which performs both training stages.

\subsection{Evaluation Metrics}
An overview of the evaluation metrics used in the experiments is presented in Fig.~\ref{fig:4}.
We varied the number of paired data of descriptions and actions, and verified the effectiveness of the proposed method from the results of two tasks: translation from action to description (Experiment 1) and translation from description to action (Experiment 2).

\begin{figure*}[tb]
\centering
\includegraphics[height=11.0cm]{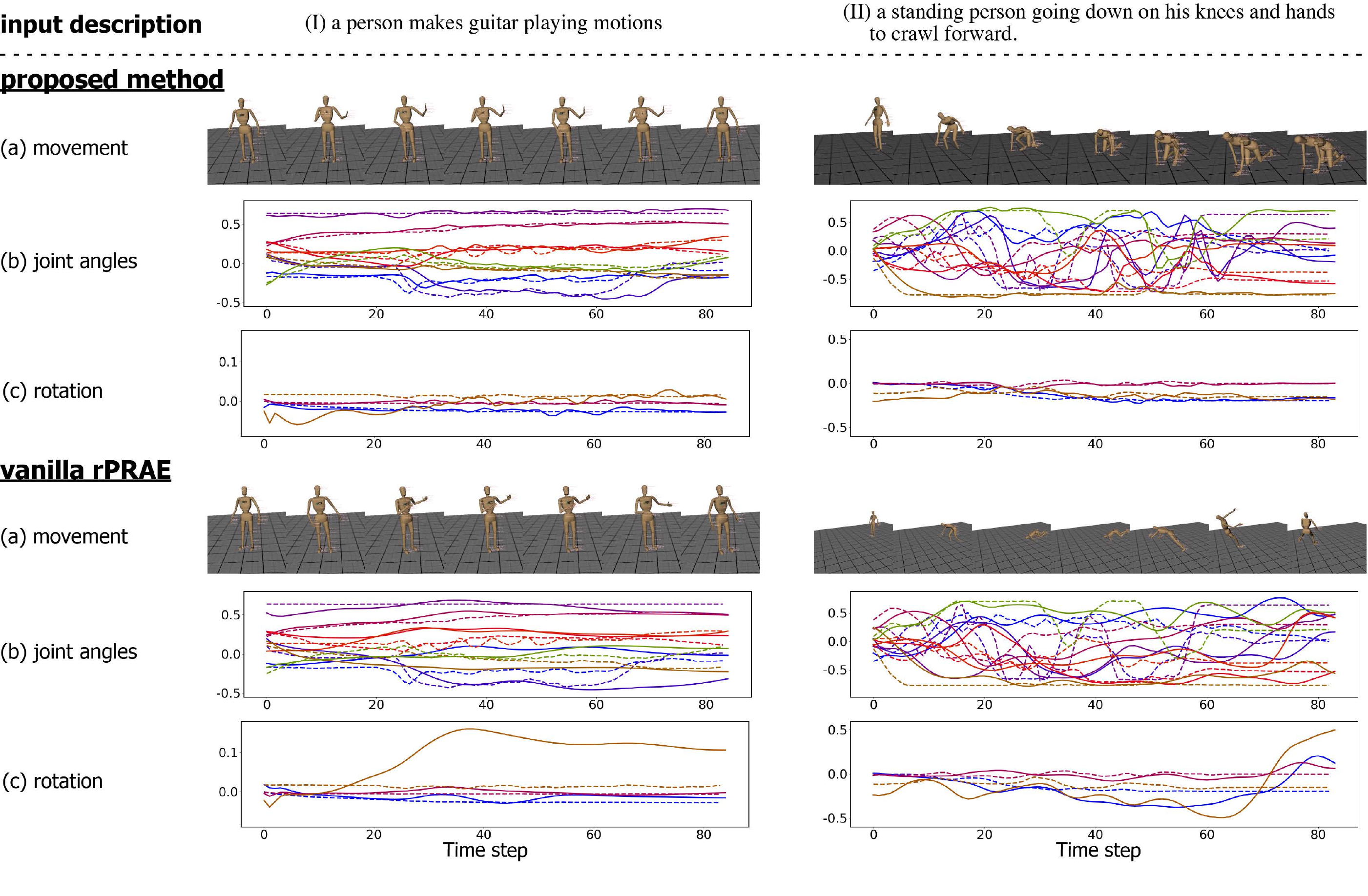}
\caption{Generated samples of actions by inputting each description: (I) and (II).
The proposed model was pre-trained with 4368 non-paired data and fine-tuned with 500 paired data, same as Fig.~\ref{fig:5-1}.
Vanilla rPRAE was also trained with 500 paired data.
(a) shows the images of the generated actions, and (b) and (c) show their joint angles and rotations, respectively. 
Positions are not shown due to space limitations, but the results are similar to (b) and (c).
The solid lines indicate the generated values. The broken lines indicate the correct trajectories. 
Among the 44 joint angles, eight of them are shown in (b).
}
\label{fig:5-2}
\end{figure*}

\begin{table}[tb]
\centering
\small
\tabcolsep 3pt
\caption{Comparision of Scores of Translated Descriptions\\by Amount of Paired Data}
\label{tab:2}
\scalebox{0.9}{
    \begin{tabular}{ll|ccccccc}
    \hline
    \ Paired & Non-Paired & BLEU & $\rm{P_{BERT}}$ & $\rm{R_{BERT}}$ & $\rm{F_{BERT}}$ & SimCSE\\
    \hline
    \hline
    4368 & 0 & .269 & .905 & .902 & .903 & .667 \\
    500  & 0 & .227 & .897 & .896 & .896 & .612 \\
    250  & 0 & .218 & .896 & .891 & .893 & .592 \\
    100  & 0 & .195 & .889 & .887 & .887 & .560 \\
    \hline
    500  & 4368 & .236 & .901 & .897 & .899 & .603 \\
    250  & 4368 & .234 & .900 & .894 & .897 & .597 \\
    100  & 4368 & .196 & .890 & .888 & .889 & .572 \\
    \hline
    \end{tabular}
}
\end{table}


\begin{table}[tb]
\centering
\small
\tabcolsep 3pt
\caption{Comparison of Scores of Translated Descriptions \\
by Amount of Non-Paired Data}
\scalebox{0.9}{
    \begin{tabular}{ll|ccccccc}
    \hline
    \ Paired & Non-Paired & BLEU & $\rm{P_{BERT}}$ & $\rm{R_{BERT}}$ & $\rm{F_{BERT}}$ & SimCSE\\
    \hline
    \hline
    500 & 3000 & .229 & .899 & .897 & .898 & .608 \\
    500 & 2000 & .230 & .899 & .896 & .898 & .606 \\
    500 & 1000 & .223 & .897 & .897 & .897 & .603 \\
    500 & 750  & .223 & .896 & .896 & .896 & .601 \\
    \hline
    \end{tabular}
}
\label{tab:3}
\end{table}


\textbf{Experiment 1: }
We quantitatively evaluated the performance of the model for description generation  using (1) multiple evaluation metrics for machine translation. In our experiments, we used  BLEU~\cite{Papineni2002_bleu}, BERTScore~\cite{bert-score}, and cosine similarity of the sentence vectors obtained from SimCSE~\cite{gao2021simcse}. BLEU is a measure that calculates the agreement between the reference sentence and included words without considering context information. BERTScore uses a vector representation of tokens, including contextual information acquired from the language model BERT~\cite{bert}; it has been reported to have a higher correlation with human ratings than BLEU. SimCSE is a model that uses BERT to embed the meaning of sentences in a more compatible way with human senses. For all of these scores, the higher the number, the closer the sentence is generated to the reference sentence.

\textbf{Experiment 2: }
In the translation from description to action, quantitative evaluation is difficult because there is no fixed scale to evaluate each motion.
Therefore, in this study, we evaluated the following points by referring to the conventional methods for bidirectional translation: 
(2-1) the back-translated sentences (description$\rightarrow$action$\rightarrow$description) using BLEU, BERTScore, and SimCSE as in Experiment 1, and
(2-2) relative performance~\cite{Plappert2018}, which is the recovery rate of the generated action.
Relative performance is the percentage of translation quality metrics that are retained after translating the description into action and then transforming this generated action back into a description using the model to be compared with.
In our experiment, we used a reference model trained with all the paired data for comparison.

\section{Results and Discussion}
\subsection{Experiment 1: Translation from Action to Description}
First, we evaluated the proposed method in translation from action to description.
The translation quality metrics of the generated descriptions are shown in Table~\ref{tab:2}.
The first and second columns show the number of paired and non-paired data used for training, respectively.
The top four rows show the conventional method (vanilla rPRAE) trained with paired data only, and the bottom three rows show the proposed method pre-trained on non-paired data and fine-tuned on paired data.
Each result shows the average of three training trials using random seeds.
In the vanilla rPRAE, all non-paired data numbers were zero because pre-training was not performed.

When comparing the methods with the same amount of paired data, it was confirmed that the proposed method produced sentences that were closer to the correct sentences than the vanilla rPRAE as the amount of paired data became smaller.
This tendency was observed regardless of the type of score, and when the amount of paired data was less than 250 (approximately 5.7\% of the total data), the proposed method outperformed the vanilla rPRAE in all cases.
However, when the amount of paired data exceeded 500 (approximately 11.4\% of the total data), the vanilla rPRAE rarely outperformed the proposed method.
This is presumably because 500 paired data are almost sufficient for the KIT dataset, and the impact of pre-training using non-paired data became smaller.

\begin{table}[tb]
\centering
\small
\tabcolsep 3pt
\caption{Comparison of Scores of Back-Translated Descriptions\\
by Amount of Paired Data}
\scalebox{0.9}{
    \begin{tabular}{ll|ccccccc}
    \hline
    \ Paired & Non-Paired & BLEU & $\rm{P_{BERT}}$ & $\rm{R_{BERT}}$ & $\rm{F_{BERT}}$ & SimCSE\\
    \hline
    \hline
    4368 & 0 & .259 & .905 & .901 & .903 & .648 \\
    500  & 0 & .228 & .900 & .898 & .899 & .600 \\
    250  & 0 & .227 & .899 & .897 & .898 & .588 \\
    100  & 0 & .217 & .899 & .896 & .897 & .577 \\
    \hline
    500  & 4368 & .242 & .903 & .899 & .901 & .618 \\
    250  & 4368 & .235 & .902 & .898 & .900 & .604 \\
    100  & 4368 & .232 & .903 & .897 & .900 & .595 \\
    \hline
    \end{tabular}
}
\label{tab:4}
\end{table}

\begin{table}[tb]
\centering
\small
\tabcolsep 3pt
\caption{Comparison of Relative Performance of Back-Translated\\ Descriptions by Amount of Paired Data [\%]}
\scalebox{0.9}{
    \begin{tabular}{ll|ccccccc}
    \hline
    \ Paired & Non-Paired & BLEU & $\rm{P_{BERT}}$ & $\rm{R_{BERT}}$ & $\rm{F_{BERT}}$ & SimCSE\\
    \hline
    \hline
    4368 & 0 & 96.4 & 100. & 99.9 & 100. & 98.8 \\
    500  & 0 & 84.8 & 99.4 & 99.6 & 99.5 & 91.4 \\
    250  & 0 & 84.5 & 99.3 & 99.5 & 99.4 & 89.7 \\
    100  & 0 & 80.7 & 99.3 & 99.4 & 99.4 & 88.0 \\
    \hline
    500  & 4368 & 90.2 & 99.8 & 99.7 & 99.7 & 94.2 \\
    250  & 4368 & 87.6 & 99.7 & 99.6 & 99.7 & 92.0 \\
    100  & 4368 & 86.3 & 99.8 & 99.5 & 99.7 & 90.6 \\
    \hline
    \end{tabular}
}
\label{tab:5}
\end{table}

To verify the effect of non-paired data on the proposed method, we investigated the scores shown in Table~\ref{tab:3}.
Except for the number of non-paired data, the training settings remained unchanged.
The overall trend was that the smaller the non-paired data, the lower the score.
When the amount of non-paired data was 750 (the fourth row of Table~\ref{tab:2}), it was confirmed that errors at the word level occurred frequently, although grammatically appropriate sentences were generated.
Therefore, the translation quality metrics can be improved by collecting non-paired data, which indicates the effectiveness of the proposed method using an independent pre-trained model.

In addition to the quantitative evaluation using translation quality metrics, we also conducted qualitative evaluations by visually checking the generated descriptions.
We confirmed appropriate description generation for actions in which the vanilla rPRAE failed to translate correctly (Figure~\ref{fig:5-1}).
This is because of the improvement of the expressive ability in the model by pre-training using non-paired data, as described earlier.
However, there were some cases in which the proposed method failed to generate the data while the vanilla rPRAE succeeded.
It is often possible to confirm that the vanilla rPRAE generates appropriate actions in cases where similar motions are included in the training data.
From these results, it can be inferred that, although the vanilla rPRAE was originally powerful in verbalizing behaviors included in the training data, the proposed learning method improved the generalization ability
as a trade-off.

\subsection{Experiment 2: Translation from Description to Action}
Next, we evaluated our method from the viewpoint of translation from description to action.
Table~\ref{tab:4} shows the results of the back-translation evaluation of the generation actions.
Table~\ref{tab:5} shows their relative performance to the reference model (first row in Table~\ref{tab:2}) as illustrated in Fig.~\ref{fig:4}.
The format of each table is the same as that of Table~\ref{tab:2}.
Note that only Table~\ref{tab:5} is presented in percentage.
In the results of each table, the proposed method was found to significantly improve the score compared with the vanilla rPRAE for any number of paired data.
In particular, the relative performance shows good results, confirming the effectiveness of the proposed method.

We investigated the scores and relative performance of the back-translated descriptions for different amount of non-paired data, as shown in Tables~\ref{tab:6} and \ref{tab:7}, respectively.
The overall score tends to decline as the amount of non-paired data decreases, suggesting that collecting non-paired data can improve the quality of the generated sentence.
In addition, the score of the proposed method was much higher than that of Experiment 1, indicating that the proposed method is particularly suitable for translation from description to action.
This is because the representations of the actions acquired by pre-training were more diverse than those of the descriptions.

\begin{table}[tb]
\centering
\small
\tabcolsep 3pt
\caption{Comparison of Scores of Back-Translated Descriptions \\
by Amount of Non-Paired Data}
\scalebox{0.9}{
    \begin{tabular}{ll|ccccccc}
    \hline
    \ Paired & Non-Paired & BLEU & $\rm{P_{BERT}}$ & $\rm{R_{BERT}}$ & $\rm{F_{BERT}}$ & SimCSE\\
    \hline
    \hline
    500 & 3000 & .247 & .903 & .899 & .901 & .618 \\
    500 & 2000 & .238 & .903 & .899 & .901 & .614 \\
    500 & 1000 & .240 & .902 & .899 & .900 & .609 \\
    500 & 750  & .231 & .901 & .899 & .900 & .611 \\
    \hline
    \end{tabular}
}
\label{tab:6}
\end{table}

\begin{table}[tb]
\centering
\small
\tabcolsep 3pt
\caption{
\vspace{-3pt}
Comparison of Relative Performance of Back-Translated Descriptions by Amount of Non-Paired Data [\%]}
\scalebox{0.9}{
    \begin{tabular}{ll|ccccccc}
    \hline
    \ Paired & Non-Paired & BLEU & $\rm{P_{BERT}}$ & $\rm{R_{BERT}}$ & $\rm{F_{BERT}}$ & SimCSE\\
    \hline
    \hline
    500 & 3000 & 92.1 & 99.8 & 99.7 & 99.7 & 94.2 \\
    500 & 2000 & 88.5 & 99.8 & 99.8 & 99.8 & 93.6 \\
    500 & 1000 & 89.5 & 99.7 & 99.7 & 99.7 & 92.8 \\
    500 & 750  & 86.1 & 99.6 & 99.7 & 99.6 & 93.1 \\
    \hline
    \end{tabular}
}
\label{tab:7}
\end{table}

\begin{figure}[t]
\centering
\includegraphics[width=\columnwidth]{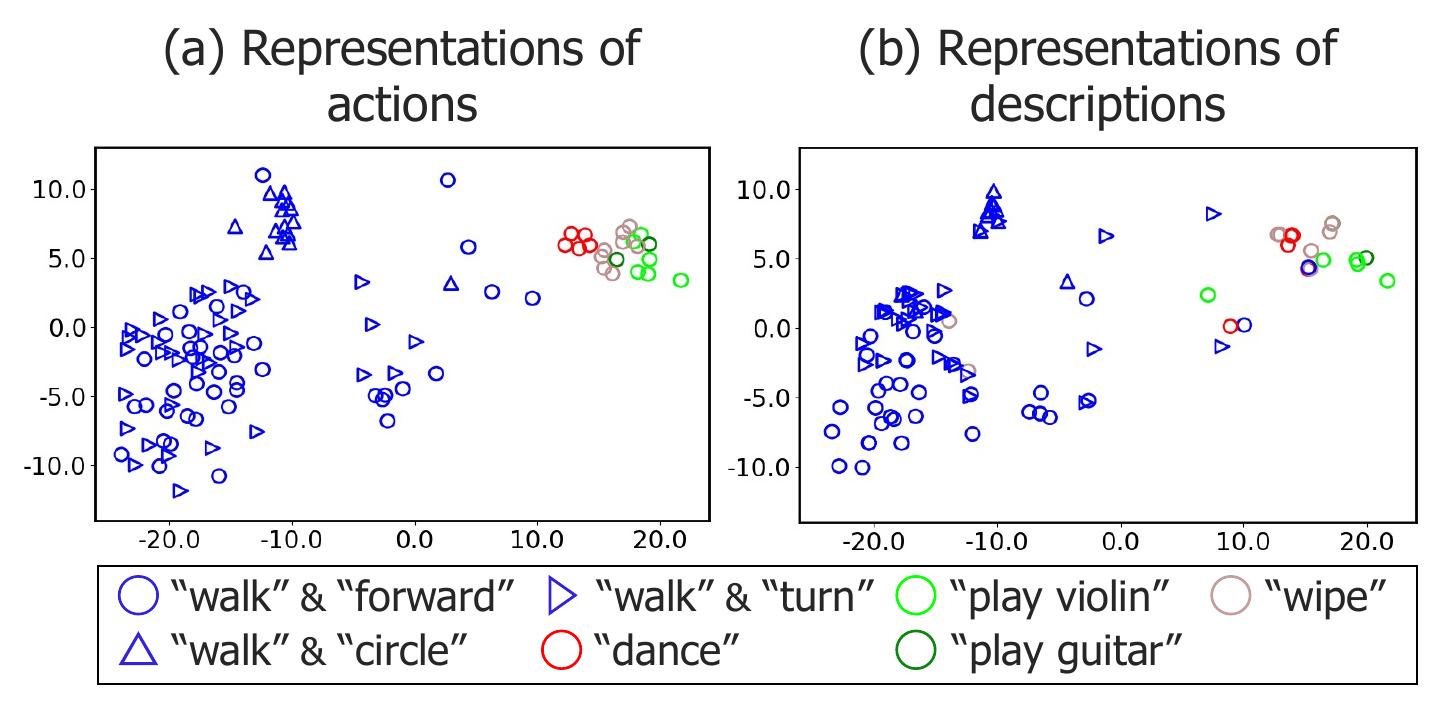}
\caption{t-SNE visualization of the shared representations.
(a) Each plot represents an action. (b) Each plot represents a description. The labels were manually assigned for each action.}
\label{fig:6}
\end{figure}

As in Experiment 1, we conducted a qualitative evaluation by visually checking the generated samples.
In most settings, the proposed method generated actions comparable to those of the reference model, whereas the vanilla rPRAE generated many cases of indistinguishable actions.
Figure~\ref{fig:5-2} showed that the proposed method repeatedly generated proper actions, although the start timings were different from the correct ones.
In summary, the proposed method can appropriately bind the intermediate representations of the description and action RAEs by fine-tuning with small paired data.
Moreover, their bidirectional transformation was realized.
This study confirms the effectiveness of the proposed two-stage training concept.
However, owing to the simplicity of the proposed method, there is room for improvement, such as the use of non-paired data from different settings
and the introduction of metric learning for clustering using positive/negative information~\cite{simclr}\cite{moco}.
The latter is considered an effective method based on the visualization results of the intermediate representations described in the next subsection.

\subsection{Comparison of Shared Representations}
Finally, to analyze the model behaviors, the intermediate representations of both the encoded descriptions and actions were visualized using t-SNE.
Figure~\ref{fig:6} (a) shows the representation $z^{\rm act}$ by the action RAE, and Fig.\ref{fig:6} (b) shows the representation $z^{\rm dsc}$ by the description RAE.
The visualized model was trained with 500 paired data and 4368 non-paired data.
The plots in the figure show the encoded actions in the test data with the labels ``walk,'' ``wipe,'' ``play,'' and ``dance,'' which are frequently occurring verbs.
In addition, samples with ``walk'' contained supplementary information about the direction of movement, such as ``forward,'' ``turn,'' and ``circle.''

Although the boundaries were mixed, the samples with the same label were encoded in a similar position in the intermediate representation of each RAE.
In particular, movements related to ``walk'' and those related to the upper body (``wipe,'' ``play,'' ``dance'') were clearly separated.
Additionally, within the actions with the ``walk'' label, only apparently distinctly different actions (``circle'') were appropriately separated.
Therefore, it was confirmed that the intermediate representations of each RAE were bound, even when fine-tuning was performed with small paired data.

Some of the generated samples have complex structures in the feature space. In many cases of ``walk''+``forward'' and ``walk''+``turn,'' they encompassed multiple motions, which were widely scattered in the feature space. These two actions were considered to overlap because the beginnings of their motions were similar. In addition, because the description RAE reflected factors other than the verb labels employed in this analysis, the samples were not clearly grouped compared to the action RAE. The results were also significantly affected by orthographical variants in the captions, particularly for the ``dance'' label samples.
Additionally, the words ``right'' and ``left'' were used in ``walk''+``turn,'' ``wipe,'' and ``play'' to indicate the direction of movement or the moving hand, which may have affected the structuring in the feature space. This may also be due to the vector representation of Word2Vec, which does not consider contextual information. 
We believe that utilizing the different properties of these intermediate representations in the generation phase is the next step toward improving the model.








\section{Conclusion}
This study proposed a learning method for bidirectional translation between actions and descriptions using small paired data.
In the proposed method, RAEs for actions and descriptions were trained on large-scale non-paired data, and then the entire model was fine-tuned on small-scale paired data.
As a result of the experiments on the mutual generation of actions and descriptions, it was confirmed that the proposed method could realize appropriate bidirectional transformation even if the paired data connecting them are small.
The visualized intermediate representations of each RAE showed that similar actions were encoded in a clustered position and the corresponding feature vectors were well aligned.
We could use any encoder--decoder model for each modality because their structures are not constrained.

The use of context-independent word embeddings, however, had the limitation that it may not be able to properly handle polysemous expressions.
We plan to improve the performance of our method by incorporating more powerful context-dependent language models, such as Transformer~\cite{transformer}.
Furthermore, because the interaction with various objects and controlling the timing to generate speech and actions are essential to realize a collaborative robot, we would like to collect data and address these issues using the robot AIREC\footnote{\url{https://airec-waseda.jp/en/toppage_en/}}.

\addtolength{\textheight}{-12cm}   


\section*{ACKNOWLEDGMENT}
This work was supported by JST [Moonshot R\&D][Grant Number JPMJMS2031].

\bibliographystyle{IEEEtran}
\bibliography{IEEEabrv,refs}

\begin{thebibliography}{10}
\providecommand{\url}[1]{#1}
\csname url@rmstyle\endcsname
\providecommand{\newblock}{\relax}
\providecommand{\bibinfo}[2]{#2}
\providecommand\BIBentrySTDinterwordspacing{\spaceskip=0pt\relax}
\providecommand\BIBentryALTinterwordstretchfactor{4}
\providecommand\BIBentryALTinterwordspacing{\spaceskip=\fontdimen2\font plus
\BIBentryALTinterwordstretchfactor\fontdimen3\font minus
  \fontdimen4\font\relax}
\providecommand\BIBforeignlanguage[2]{{%
\expandafter\ifx\csname l@#1\endcsname\relax
\typeout{** WARNING: IEEEtran.bst: No hyphenation pattern has been}%
\typeout{** loaded for the language `#1'. Using the pattern for}%
\typeout{** the default language instead.}%
\else
\language=\csname l@#1\endcsname
\fi
#2}}

\bibitem{Yamada2018_PRAE}
T.~Yamada, H.~Matsunaga, and T.~Ogata, ``Paired recurrent autoencoders for
  bidirectional translation between robot actions and linguistic
  descriptions,'' \emph{IEEE Robotics and Automation Letters}, vol.~3, no.~4,
  pp. 3441--3448, 2018.

\bibitem{Toyoda2021_rPRAE}
M.~Toyoda, K.~Suzuki, H.~Mori, Y.~Hayashi, and T.~Ogata, ``Embodying
  pre-trained word embeddings through robot actions,'' \emph{IEEE Robotics and
  Automation Letters}, vol.~6, no.~2, pp. 4225--4232, 2021.

\bibitem{Matthews2019}
D.~Matthews, S.~Kriegman, C.~Cappelle, and J.~Bongard, ``Word2vec to behavior:
  morphology facilitates the grounding of language in machines,'' in
  \emph{Proceedings of the 2019 IEEE/RSJ International Conference on
  Intelligent Robots and Systems}, 2019, pp. 4153--4160.

\bibitem{Qiu2020pre-train}
X.~Qiu, T.~Sun, Y.~Xu, Y.~Shao, N.~Dai, and X.~Huang, ``Pre-trained models for
  natural language processing: A survey,'' \emph{Science China Technological
  Sciences}, vol.~63, no.~10, pp. 1872--1897, Sep 2020.

\bibitem{zellers2021piglet}
R.~Zellers, A.~Holtzman, M.~Peters, R.~Mottaghi, A.~Kembhavi, A.~Farhadi, and
  Y.~Choi, ``Piglet: Language grounding through neuro-symbolic interaction in a
  3d world,'' in \emph{Proceedings of the 59th Annual Meeting of the
  Association for Computational Linguistics}, 2021, pp. 2040--2050.

\bibitem{ruan2022video-language}
L.~Ruan and Q.~Jin, ``Survey: Transformer based video-language pre-training,''
  \emph{AI Open}, vol.~3, pp. 1--13, 2022.

\bibitem{bert}
J.~Devlin, M.-W. Chang, K.~Lee, and K.~Toutanova, ``Bert: Pre training of deep
  bidirectional transformers for language understanding,'' in \emph{Proceedings
  of the 2019 Conference of the North American Chapter of the Association for
  Computational Linguistics: Human Language Technologies, Volume 1}, 2019, pp.
  4171--4186.

\bibitem{huang2021unifying}
Y.~Huang, B.~Liu, and Y.~Lu, ``Unifying multimodal transformer for
  bi-directional image and text generation,'' in \emph{Proceedings of the 29th
  ACM International Conference on Multimedia}, 2021.

\bibitem{baltruvsaitis2018multimodal}
T.~Baltru{\v{s}}aitis, C.~Ahuja, and L.-P. Morency, ``Multimodal machine
  learning: A survey and taxonomy,'' \emph{IEEE transactions on pattern
  analysis and machine intelligence}, vol.~41, no.~2, pp. 423--443, 2018.

\bibitem{NODA2014721}
K.~Noda, H.~Arie, Y.~Suga, and T.~Ogata, ``Multimodal integration learning of
  robot behavior using deep neural networks,'' \emph{IEEE Robotics and
  Autonomous Systems}, vol.~62, no.~6, pp. 721--736, 2014.

\bibitem{hasegawa-etal-2017-incorporating}
M.~Hasegawa, T.~Kobayashi, and Y.~Hayashi, ``Incorporating visual features into
  word embeddings: A bimodal autoencoder-based approach,'' in \emph{Proceedings
  of the 12th International Conference on Computational Semantics}, 2017.

\bibitem{felbo2017using}
B.~Felbo, A.~Mislove, A.~S{\o}gaard, I.~Rahwan, and S.~Lehmann, ``Using
  millions of emoji occurrences to learn any-domain representations for
  detecting sentiment, emotion and sarcasm,'' in \emph{Proceedings of the 2017
  Confefence on Empirical Methods in Natural Language Processing}, 2017, pp.
  1615--1625.

\bibitem{Plappert2016_KIT}
M.~Plappert, C.~Mandery, and T.~Asfour, ``The {KIT} motion-language dataset,''
  \emph{Big Data}, vol.~4, no.~4, pp. 236--252, 2016.

\bibitem{Plappert2018}
------, ``Learning a bidirectional mapping between human whole-body motion and
  natural language using deep recurrent neural networks,'' \emph{IEEE Robotics
  and Autonomous Systems}, vol. 109, pp. 13--26, 2018.

\bibitem{Mikolov2013_w2v}
T.~Mikolov, I.~Sutskever, K.~Chen, G.~S. Corrado, and J.~Dean, ``Distributed
  representations of words and phrases and their compositionality,''
  \emph{Advances in neural information processing systems}, vol.~26, 2013.

\bibitem{bi-rnn}
M.~Schuster and K.~K. Paliwal, ``Bidirectional recurrent neural networks,''
  \emph{IEEE transactions on Signal Processing}, vol.~45, no.~11, pp.
  2673--2681, 1997.

\bibitem{adam}
D.~P. Kingma and J.~Ba, ``Adam: A method for stochastic optimization,''
  \emph{arXiv preprint arXiv:1412.6980}, 2014.

\bibitem{Papineni2002_bleu}
K.~Papineni, S.~Roukos, T.~Ward, and W.-J. Zhu, ``Bleu: a method for automatic
  evaluation of machine translation,'' in \emph{Proceedings of the 40th annual
  meeting of the Association for Computational Linguistics}, 2002, pp.
  311--318.

\bibitem{bert-score}
T.~Zhang*, V.~Kishore*, F.~Wu*, K.~Q. Weinberger, and Y.~Artzi, ``Bertscore:
  Evaluating text generation with bert,'' in \emph{Proceedings of the
  International Conference on Learning Representations}, 2020.

\bibitem{gao2021simcse}
T.~Gao, X.~Yao, and D.~Chen, ``{S}im{CSE}: Simple contrastive learning of
  sentence embeddings,'' in \emph{Proceedings of the 2021 Conference on
  Empirical Methods in Natural Language Processing}, 2021, pp. 6894--6910.

\bibitem{simclr}
T.~Chen, S.~Kornblith, K.~Swersky, M.~Norouzi, and G.~Hinton, ``Big
  self-supervised models are strong semi-supervised learners,'' \emph{arXiv
  preprint arXiv:2006.10029}, 2020.

\bibitem{moco}
X.~Chen, H.~Fan, R.~Girshick, and K.~He, ``Improved baselines with momentum
  contrastive learning,'' \emph{arXiv preprint arXiv:2003.04297}, 2020.

\bibitem{transformer}
A.~Vaswani, N.~Shazeer, N.~Parmar, J.~Uszkoreit, L.~Jones, A.~N. Gomez, L.~u.
  Kaiser, and I.~Polosukhin, ``Attention is all you need,'' in \emph{Advances
  in Neural Information Processing Systems}, vol.~30, 2017.

\end{thebibliography}

\end{document}